\documentclass[journal]{IEEEtran}
\usepackage{graphicx}
\usepackage{algorithm,algorithmic}
\usepackage{amsmath,amssymb,amsfonts}
\usepackage{float}
\usepackage{enumerate}
\usepackage{stfloats}
\usepackage{color}
\usepackage{url}

\makeatletter
\def\url@leostyle{%
  \@ifundefined{selectfont}{\def\UrlFont{\sf}}{\def\UrlFont{\small\ttfamily}}}
\makeatother
\ifCLASSINFOpdf
\else
\fi

\begin{document}

\title{Collaborative City Digital Twin For Covid-19 Pandemic: A Federated Learning Solution}

%
% author names and IEEE memberships
% note positions of commas and nonbreaking spaces ( ~ ) LaTeX will not break
% a structure at a ~ so this keeps an author's name from being broken across
% two lines.
% use \thanks{} to gain access to the first footnote area
% a separate \thanks must be used for each paragraph as LaTeX2e's \thanks
% was not built to handle multiple paragraphs
%

\author{Junjie Pang,
        Jianbo Li,
        Zhenzhen Xie,
        Yan Huang,
        and~Zhipeng Cai,~\IEEEmembership{Senior Member,~IEEE}% <-this % stops a space
\thanks{This work is partly supported by the U.S. National Science Foundation (NSF) under grant NOs. 1741277, 1829674,1704287, and 1912753, in part by the National Key Research and Development Plan of China under Grant 2017YFA0604500, in part by the National Natural Science Foundation of China under Grant 61701190, in part by the Key Technology Innovation Cooperation Project of Government and University for the whole Industry Demonstration under Grant SXGJSF2017-4, the Program for Innovative Postdoctoral Talents in Shandong Province under Grant 40618030001, and National Key Research and Development Program of China under Grant 2018YFB2100303}% <-this % stops a space
\thanks{Junjie Pang is with the Business School and the College of Computer Science and Technology, Qingdao University, Qingdao, Shandong 266000, China (e-mail: pangjj18@163.com).}% <-this % stops a space
\thanks{Jianbo Li is with the College of Computer Science and Technology, Qingdao University, Qingdao, Shandong 266000, China (e-mail: lijianbo@188.com).}% <-this % stops a space
\thanks{Zhenzhen Xie is with the College of Computer Science and Technology, Jilin University, Changchun, Jilin 130012, China (e-mail: xiezz14@mails.jlu.edu.cn).}% <-this % stops a space
\thanks{Yan Huang is with the College of Computing and Software Engineering  at  Kennesaw  State  University, Atlanta, GA, USA (e-mail: yhuang24@kennesaw.com).}% <-this % stops a space
\thanks{Zhipeng Cai is with the Department of Computer Science, Georgia State University, Atlanta, GA, USA (e-mail: zcai@gsu.edu).}
\thanks{Corresponding Author: Yan Huang}% <-this % stops a space

}

% The paper headers
\markboth{Journal of \LaTeX\ Class Files,~Vol.~14, No.~8, August~2015}%
{Shell \MakeLowercase{\textit{et al.}}: Bare Demo of IEEEtran.cls for IEEE Journals}

\maketitle

\begin{abstract}
The novel coronavirus, COVID-19, is a crisis that affects all segments of the population.
As the knowledge and understanding of COVID-19 evolve, an appropriate response plan for this pandemic is regarded as the most effective method to control the spreading.
Recent studies indicate that the city digital twin is beneficial to tackle this health crisis since it can construct a virtual replica to simulate factors such as climate conditions, response operations, and people's trajectory to help plan efficient and inclusive decisions.
However, the city digital twin system relies on long-term and high-quality data collection to make appropriate decisions, which limited its advantages when facing urgent crisis like COVID-19.
Federated Learning (FL), where all the clients can learn a shared model while keeping all the training data locally, emerges as a promising solution to accumulate the insights from multiple data source efficiently.
Furthermore, the enhanced privacy protection settings break the privacy barriers lies in such collaboration.
In this work, we propose a collaborative city digital twin based on FL, a novel paradigm that allowing multiple city DT to share the local strategy and status in a timely manner.
In particular, an FL central server manages the local updates of multiple collaborators (city DT), provides a global model which is trained in multiple iterations at different city DT systems, until the model gains the correlations between various response plan and infection trend.
That means, a collaborative city DT paradigm based on FL techniques can obtain knowledge and patterns from multiple DTs, and eventually establish a `global view' for city crisis management.
Meanwhile, it also helps to improve each city digital twin selves by consolidating other DT's respective data without violating privacy rules.
To validate the proposed solution, we take COVID-19 pandemic as a case study.
The experimental results on the real dataset with various response plan validate our proposed solution and demonstrate the superior performance.
\end{abstract}

% Note that keywords are not normally used for peerreview papers.

\IEEEpeerreviewmaketitle

\section{Introduction}

\IEEEPARstart{C}OVID-19, an infectious disease caused by the most recently discovered coronavirus was identified in 2019\cite{WHO}, December 31th\footnote{https://www.who.int/emergencies/diseases/novel-coronavirus-2019}.
The virus has spread worldwide in less than three months, but infected more than 5.7 million people around the world and caused over 844,000 deaths\footnote{https://www.worldometers.info/coronavirus/}.
This novel coronavirus outbreak leads to severe illness and deaths in both younger and middle-aged adults, and it also critically affect people's daily lives.
For example, almost 500 million schoolchildren have been cut off from learning recently for avoiding direct exposure to this disease; thousands of people canceled their trips to reduce the risks of potential infection, etc.
So far, there is still no specific antiviral treatment for COVID-19.
Fortunately, we observe that various active response plans have a clear positive effect upon preventing transmission from one person to another.
One of the successful examples is in South Korea\footnote{https://ourworldindata.org/covid-exemplar-south-korea}. Vt taking early, aggressive government response including establishing a large number of detection sites, isolating the infected patients in a very short time, and tracing contacts for searching the potential infected people with thoroughness, South Korea controlled the spreading efficiently.
Thus, we observed that South Korea has quickly flattened the epidemic curve after its first peak in February 2020, and the daily cases declined to nearly zero after two months.
The similar observation can also be found in many other countries, such as China, Germany and the United Kingdom.
That means, if we response appropriately, the infection spreading can be effectively controlled.
Consequently, it is necessary to establish a paradigm to collect the historical data, and distinguish the effectiveness of various response plans.
We also expect that such a paradigm could be used for other urgent city crisis.
%**************************************************************************************
To obtain the above goal, city Digital Twin (city DT) has demonstrated a remarkable potential to offer the function of data collection, in-depth analysis and most important, it operationalizes real-world insights to asset performance and refine future actions~\cite{DBLP:journals/jms/CroattiGMR20},~\cite{DBLP:books/sp/20/BagariaLBAVE20}.
In particular, a city digital twin is a virtual replica of the city that can provide digital view of city facilities, human activities and behavior, and other types of information about urban area.
By keep interacting with the target environment, a city digital twin not only could automatically learn the patterns and generate useful insights from the historical data, but also continuously improve itself by verifying the automatic operations within the real world. 
Eventually, a city DT enables interdisciplinary convergence in multiple aspects of city informatics, and offers increased visibility and machine intelligence for both human assisted decisions and automatic operations to facilitate city government and planners' decision-making.
Several successful attempts, including Columbus Consolidated Government (Columbus), Digital Twin City of Atlanta, and Virtual Singapore, validate city digital twin's value for capture the complex spatial and temporal implications among multiple data source to optimize urban sustainability on traffic management, architectural planning and energy saving solutions.
Meanwhile,~\cite{fan2019disaster} proposed a novel framework of disaster city digital twin with deep learning techniques, which presents the visions for a city DT based disaster management and further explains the advantages to response the disaster and crisis.
Thus, we expect that the city DT could be a promising solution to tackle current health crisis by simulating various response plan in this virtual system, and finally determine the proper actions to control the infection trend.

%**************************************************************************************
Note that, when we aim to construct a city digital twin to model COVID-19 pandemic and generate the prediction of a specified response plan, we need to resolve the data sparsity challenge with considerations of privacy by following reasons.
First, a city digital twin for health crisis like COVID-19 may contain more sensitive information than other application scenarios, which imply the pressing need for privacy protection on the raw data.
Second, as each city constructing its own digital twin system, which means it could form an isolated data island.
Under such conditions, each city can only depend on a trail-error manner to control the pandemic instead of learning experience from others, which may result in delayed response or inappropriate operations.
To resolve such problems, a collaborative city digital twin paradigm can be beneficial to alleviate the data sparsity issue.
Meanwhile, it is also helpful on pursuing an accurate prediction for choosing a better response plan.
To this definitive end, we need to answer three primary questions:
(1) How to find a collaborative plan that can enhance the prediction results while not disturbing the interactions between city digital twin systems and the physical environment?
(2) How to ensure the collaborative solution can automatically generate different insights to adapt to different city digital twin system? 
(3) How to provide the collaborative city digital twin paradigm with privacy protection guarantee?
A straightforward solution is the distributed machine learning method with privacy protection guarantee, so that various city digital twin can upload their raw data to the central server to achieve a learning result based on data fusion results.
However, the initiative data pre-processing stage of this plan could bring extra time expenses, and the centralized training manner is likely to have higher risks on privacy leakage and security issues.

\begin{figure*}
	\centering
	\includegraphics[width=0.65\textwidth]{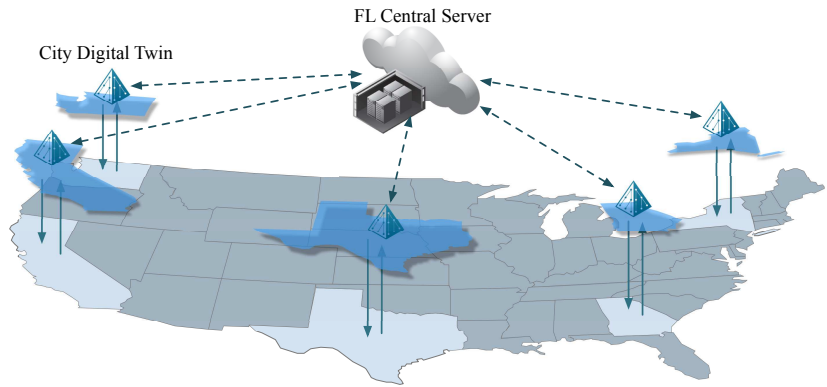}
	\setlength{\abovecaptionskip}{-0.3cm}
	\caption{Overview of the Collaborative Framework for Multiple City Digital Twin }
	\label{fig:framework}
\end{figure*}

To address the above challenges, in this paper, we propose a federated learning (FL)~\cite{DBLP:conf/aistats/McMahanMRHA17} based solution to bridge these separated city digital twin for COVID-19 pandemic control, which is depicted in~Fig.~\ref{fig:framework}.
In particular, each city digital twin is a client in our FL paradigm.
Each client uses a Sequence2sequence (Seq2seq) model to generate the local forecasting model by historical conditions of COVID-19 in targeted area.
Multiple city DTs can cooperatively train a shared model on FL platform, only with local updates without exchanging the raw data.
That is, each city DT can be improved by both local data and external insights from other city DTs to make more efficient response operations.
Our contributions can be summarized as follows.
\begin{itemize}
    \item To resolve the data sparsity challenge for COVID-19 pandemic management, we are among the first to propose a novel collaborative training framework for multiple city digital twin. 
    \item Considering the various privacy requirement among different city DT, we utilize the Federated Learning techniques as our primary solution, which offers a parameter-sharing only concept for not disturbing each city DT's privacy rules.
    \item Extensive simulations with a real dataset reveal that the our proposed framework significantly improves the performance when compared with the single DT system.
\end{itemize}

The rest of the paper is organized as follows. Section \ref{related} introduces related works.
The basic definitions and problem statement is presented in Section \ref{preliminary}.
Section \ref{fl} explains the detailed structure of the proposed framwork.
The experiments and results are analyzed in Section \ref{simulation}.
Finally, conclusion and future work are presented in Section \ref{conclusion}.

\section{Related Works}
\label{related}
Historical insights for understanding the character of different epidemics and the prediction of spreading trends have been crucially crucial for epidemic control and prevention.
Several past studies applied artificial intelligence based models to these highly specialized domains and observed impressive multi-functional properties, such as the deep neural network based short-term and high-resolution epidemic forecasting for influenza-like illness~\cite{DBLP:conf/aaai/WangCM19},
the semi-supervised deep learning framework that integrates computational epidemiology and social media mining techniques for epidemic simulation called SimNest~\cite{DBLP:conf/icdm/ZhaoCCWLR15} and EpiRP~\cite{DBLP:conf/webi/ShiZBQ019}
that using representational learning methods to capture the dynamic characteristics of epidemic spreading on social networks for epidemics-oriented clustering and classification.
To appreciate why deep learning techniques are required, in this section, we briefly review several recent relevant works and their contributions in artificial intelligence-based urban epidemic control.

\textbf{Deep Learning based Epidemic Control}:
For the past decade, deep learning technique has been used in a variety of smart health areas, including disease prevention and monitoring, intelligent diagnosis and treatment systems, and health decision-making.
Recent breakthroughs in various disease modeling, forecasting, and real-time disease surveillance have convinced us that those applications mitigate the effects of disease outbreaks.
Given various application scenarios and objectives, deep learning based models can be different:
A typical solution for localized flu nowcasting and flu activity inferring is ARGONet~\cite{lu2019improved},
which is a network-based approach leveraging spatio-temporal correlations across different states to improve the prediction accuracy.
Based on the information from influenza-related Google search frequencies, electronic health records, and historical influenza trends, ARGONet uses a spatial network to capture the spatio-temporal correlations across different states and produces more precise retrospective estimates.
Instead of leveraging multiple data source like ARGONet,~\cite{DBLP:conf/www/ZouLC18} proposes a multi-task learning based model that only using user-generated content (Web search data).
They investigate both linear and nonlinear models' capability and find they can improve the disease rate estimates significantly in the case study of influenza-like illness.
These successful attempts, however, are based on large scale data sources or plenty historical information of the disease with similar spreading patterns.
That means, high-dimensionality, irregularity forms, noise, or sparsity problems are possible to affect these learning-based models' performance. 
Such problems would be more serious when the smart health system needs to deal with unexpected infectious disease outbreaks like COVID-19, due to limited historical data and information.

\textbf{City Digital Twin}:
For most of deep learning based applications, the performance depends on the interaction with the physical world to generate data with high quality.
To fill the data gap, digital twin is proposed as a virtual representation of a device or a specific application scenario, which can interact with the target environment to collect data continuously for real-time decision making.
Such a concept leads to the advancements in both urban computing paradigm and IoT applications direct extending to smart city area recently, which termed as `city digital twin'.
It is a digital model of the physical urban environment, which collects information by hundreds of IoT devices or public databases. It provides remote decision making and real-time insights about the physical asset’s performance by utilizing machine learning and artificial intelligence techniques.
It also serves as an essential role in addressing public health, urban planning, and environmental issues.
The data streaming in the city digital twin model enables proactive remote monitoring to identify patterns and risks that foreshadow specific impending issues.
Several successful research attempts including disaster city digital twin~\cite{fan2020social,fan2019disaster}, energy management~\cite{francisco2020smart}, and city-scale Light Detection and Ranging (LiDAR) point clouds~\cite{xue2020lidar}.
Furthermore, Singapore~\cite{holstein2018virtual} and Germany~\cite{dembski2020urban} has launched the city-scale digital twin to monitor and improve utilities, which enhance the transparency, sustainability and availability of DT.

\textbf{Federated Learning}:
Pandemic control and response are the significant challenges in the field of urban computing.
However, recent research on digital twin is designed for the city resource management for a single urban area, and such application scenario often has enough historical data and multiple resources.
Unfortunately, the data resource could be the most challenging requirement for DT aiming at pandemic control, especially the unexpected contagious disease like COVID-19 due to lacking enough experience and information.
To maintain the advantages of DT and tolerate the data sparsity challenge, federated learning that allows multiple stack-holders sharing data and training global model has become a preferred scheme.
In the typical FL scheme settings, each data owner (FL client) engages in a collaborative training process without transferring the raw data to the others.
Through FL, the central server manages each client's local training updates and aggregates their contributions to enhance the global model's performance.
Several concrete scenarios include Google's Gboard, health AI and smart banking showing the advantages of federated learning in handling collaborative training issues and data difficulties among diverse data owners.
Therefore, we are motivated to utilize federated learning techniques to resolve the data sparsity challenges and design a collaborative city digital twin for COVID-19 pandemic control.

\section{Preliminary}
\label{preliminary}
In this section, we propose our designed system model of city digital twin for Covid-19 Pandemic Control. 
We first introduce the basic definitions, and then give the problem statement and explain the implementation difficulties.
\subsection{System Model}
A city digital twin has three primary components: the physical environment of the city, a virtual replica describing the city‘s architecture, functions, behaviors, and active communications between the two for obtaining real-time spatio-temporal data from various infrastructure and human systems~\cite{DBLP:conf/ssci/MohammadiT17}.
In our work, we follow this typical architecture, and propose the city digital twin for COVID-19 pandemic control using the following metrics: 

\textbf{COVID-19 infected number}: The COVID-19 infected number is the direct evidence to represent the infection status of a specific geographic area.
In our proposed DT model, the infected number in the target area is represented by: area, time, number.

\textbf{Response plan}: For COVID-19 pandemic control, various organizations and governments develop several local level or even a country-level operational plan to prepare for and respond to COVID-19. 
In our DT model, we use $R_{i} = (l_{i}, t_{st}, t_{end})$ to represent a response plan, where $l_{i}$ is the location, with $t_{st}$, $t_{end}$ denotes the starting time and end time of $R_{i}$.
The response plan includes 14 day quarantine, domestic travel limitations, gatherings limits and stay at home, nonessential business closures, reopening plans, mask policy, etc.
The effectiveness of different response plan can be varied, since it is may be affected by several external factor, such as a crowd gathers, adverse weather conditions or using vaccines.

\textbf{Temporal effects}:
For our proposed DT system, the primary goal is to analyze the specific response plan, determine whether the historical plan will impact the future infection trend, to estimate the possible plan effects.
In the process of building such a virtual replica for the COVID-19 pandemic control, we observe the temporal effects is a critical factor when we aim to provide useful guidance on response plan selection.
These temporal effects can be considered two types in our proposed DT system: temporal effects of historical infection numbers and selected response plans.

\begin{figure}
	\centering
	\includegraphics[width=0.45\textwidth]{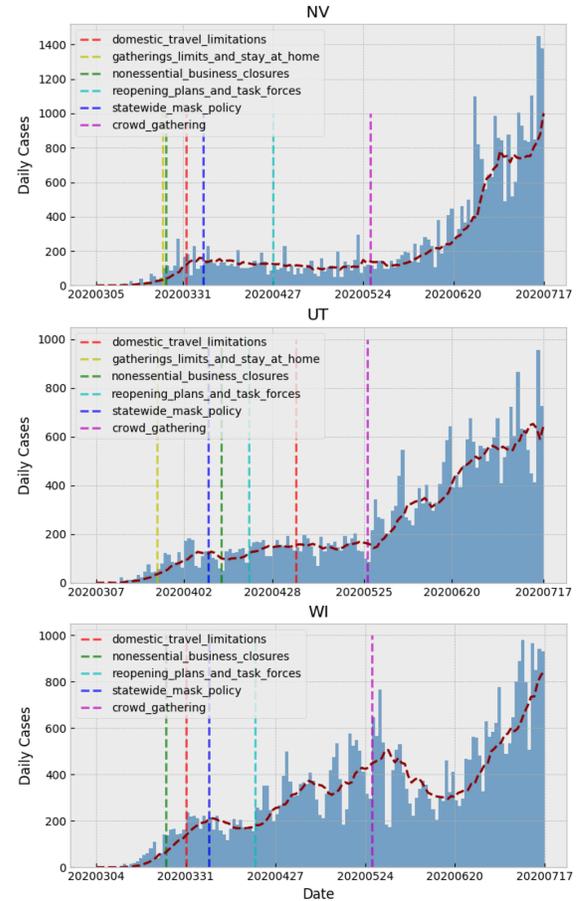}
	\setlength{\abovecaptionskip}{-0.3cm}
	\caption{Correlation between the Current Infection Trend and Historical Infection Numbers}
	\label{fig:corr}
\end{figure}

\textit{Temporal effects of historical infection numbers}: 
The historical infection numbers are direct evidence to describe the correlation between past conditions and current infection status.
In Fig~\ref{fig:corr}, we take historical daily case information of three states (NV: Nevada, UT: Utah, WI: Wisconsin) as examples to illustrate the temporal effects of multiple factors. 
From the data of early March from three states, we observe an immediate impact of historical infection numbers since it leads to a continuously increasing number of infections until April 2nd, 2020.
Such correlations play quite an essential role in explaining and predicting future infection trends.

\textit{Temporal effects of Response Plans}:
The choice of a specified response operation can also significantly impacts the number of infection.
This impact could be immediate or delayed, which may depend on the strictness of that operation or people's acceptance.
For example, in Fig.~\ref{fig:corr}, we observe that after taking a specified response plan, such as domestic travel limitations or gathering limits, the infection trend is decreased in the next several days.
Such impact could be complicated, since the 14-day time window, the time that a response plan starts to take effect, and paroxysmal public crisis may also lead to some impact on infection trends.

\subsection{Problem Statement}
With the definite goal of making the COVID-19 pandemic under control, different local agencies in each city/region may choose their own strategy to meet their local requirements.
Such divergence is mainly caused by the fact that different cities could have other conditions or expectations so that the local agency tends to provide a different response plan considering several local intrinsic properties.
For instance, area $A$, which is a thinly populated district with very low infection rates, would prefer to choose less radical response plan; while the another area $B$, where has more serious infection conditions, is very likely to choose a more strict response plan like restricting activities and closing facilities. 
Obviously, such unsynchronized operations present obstacles to accumulating knowledge about providing accurate response guidance, since each city forms an isolated data island.
Furthermore, COVID-19, as an infectious disease with unknown cause, has exaggerated the situation due to lacking enough historical information and experience for dealing with the new virus. 
That means, each city can only obtain knowledge by trial-and-error operation schemes for seeking an effective response plan, and the increasing time cost may lead to a delayed policy with poor performance.
To resolve the above difficulties, we formalize this problem as a FL training problem using the FL approach to resolve the data sparsity challenge in city digital twin construction.

Thus, we propose to learning a objective function as follows: 
Given multiple city DTs $D = {D_{1}, D_{2}, \cdots, D_{i} }$ that expecting collaborations and bounded with a local data sensing method to generate individualize data source $s_{i, 1}, s_{i, 2}, \cdots, s_{i, m_i}$, our federated training problem is aim to optimize following function:
$$
\min _{\mathbf{w}}\left\{F(\mathbf{w}) \triangleq \sum_{i=1}^{N} p_{i} F_{i}(\mathbf{w})\right\},
$$
where $p_{i} = \frac{m_{i}}{m}$ and the local objective $F_{i}(\cdot)$ is defined by
$$
F_{i}(\mathbf{w}) \triangleq \frac{1}{m_{i}} \sum_{j=1}^{m_{i}} {f} \left(\mathbf{w} ; s_{i, j}\right).
$$

\section{Exploring the Collaborative Framework For City Digital Twin}
\label{fl}
In this section, we present our proposed method utilizing FL as the collaborative platform for predicting a response plan's effectiveness, with considerations of temporal patterns and external factor impacts. 
Specifically, we describe the baseline model using Seq2seq structure, and give the details of the proposed federated training process to help understand how every single city DT can be automatically improved by such collaboration. 

A typical city digital twin actively collects real-time data from infrastructure and human systems to obtain automatic decision making or possible future behavior predictions.
Hence, it is beneficial for tracking the progress of the infectious disease in real-time and accumulating information and knowledge at city scale.
However, COVID-19 is a novel coronavirus with incredible spreading speed, which means the city DT faces the challenge of making effective operations with very limited response time. 
Under such circumstances, single city DT likely lacks experience to determine a well strategic preparedness and response plan quickly since each city DT has limited number of actions and observations.
Thus, we introduce FL as the primary solution to implement a collaborative training process, enabling multiple city DTs to train a global model for response operation prediction collaboratively. 

Federated Learning is a novel distributed learning framework with a privacy protection guarantee.
Unlike previous research dealing with training data in a centralized manner, FL uses a parameter only concept to avoid disturbing each collaborator's privacy rules. 
Thus, multiple collaborators can learn a common model as partners while preserving personal data on their own devices. 
In particular, for a new FL training task, (1) the FL central server would train a global model for initialization, then distributed this model to the existing collaborators (clients); (2) after receiving the global model, each collaborator use the local dataset to update the local parameter and generates the local updates; (3) based on specified synchronization settings, all these updates are back to the FL central server for aggregation, and the global model is improved; (4) such distributed updates iteration are repeated until the global model converges or achieve expected performance. 

From the above facts, we observe clear advantages of utilizing FL for establishing the collaborative city DTs framework: (i) by the separation of local model training and global model updates, FL offers a strong capability to deal with the isolated data island problem between multiple DTs; (ii) with enhanced privacy settings, each city DT can obtain the collaboration achievements without violating its privacy rules.
  
\subsection{System Description}
\begin{figure}
	\centering
	\includegraphics[width=.48\textwidth]{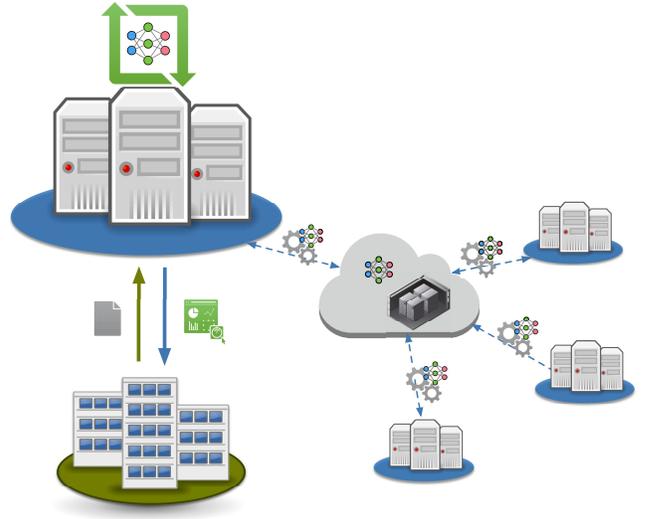}
	\caption{Long-term data updates with influence of services}
	\label{fig:system}
\end{figure}
In our proposed collaborative framework for multiple city DTs (Fig.~\ref{fig:system}), each single city DT system can reinforce itself by two ways: (1) self-renewal mode by local updates: this mode is similar as the working updates of traditional digital twin system, which collecting real-time COVID-19 conditions (historical infection numbers, external factors like crowd gathering, population age and vaccination) actively through on-device sensing, or using data source from various institutions like public health agency or hospital.
(2) joint-renewal mode by global updates: our proposed FL framework offers a platform for all City DTs with willingness for sharing its experience and insights. By such collaboration, the single city DT can gather information from others to update the local model, which we regard as a joint-renewal mode in this paper.
Specifically, our works aim to simulate a response operation's performance for guiding choosing an appropriate response plan of COVID-19.

Under such a practical context, the joint-renewal mode can effectively complete the local models by utilizing other collaborator's updates.
For example, suppose a City DT $A$ decides to take a new response operation $R_i$ to deal with a crowd-gathering caused infection outbreak; at that time, City DT $B$, which have used the same response operation $R_i$ before, can broadcast its experience through the FL platform by uploading the local updates on time.
Then, after obtaining the new global model, City DT $A$ can simulate the feedback (like trends in numbers of future infections) of response operation $R_i$ to decide if this operation is necessary.
Thus, under the FL based collaborative scheme, multiple City DTs can share the historical experience of different practical conditions by uploading the local parameter updates.
Meanwhile, the inherent character of City DT can ensure the high quality of these local updates, since each City DT is bounded with a real-time data collection basis.

\subsection{The Baseline Model: Local City Digital Twin}
\begin{figure*}
	\centering
	\includegraphics[width=0.65\textwidth]{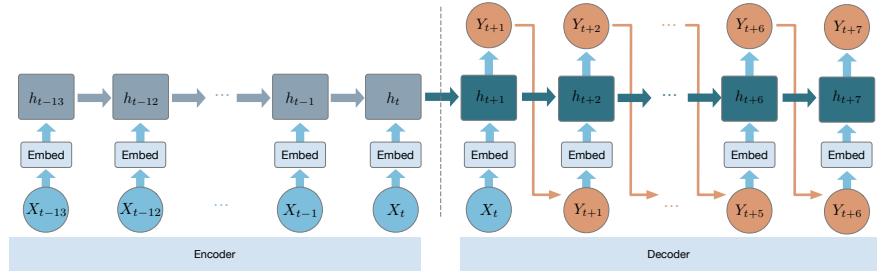}
	\setlength{\abovecaptionskip}{-0.3cm}
	\caption{Local City Diy Digital Twin model: Structure of GRU-based Seq2seq model}
	\label{fig:localDT}
\end{figure*}

For the local City DT system, we utilize the GRU based Seq2seq learning model with Gated Recurrent Unit (GRU) for local training, which is illustrated by Fig.~\ref{fig:localDT}.
The Seq2seq learning pipeline is one of the typical Encoder-Decoder structure to solve the general sequence to sequence learning problem with neural networks, which has well performance when input and the output sequences have unfixed lengths with complicated and non-monotonic relationship~\cite{DBLP:conf/nips/SutskeverVL14}.
The GRU layer is a variant of Recurrent Neural Networks (RNN) for better time-series data processing, and each GRU cell only uses two data gates to control the data from processor: the reset gate $r$ determine the information needs forget, and update gate $z$ decides how the information needs to be updated.
In particular, our proposed local City DT system using three-layer GRU for the encoder part, which maps the history of sequences (the infection number of the target area and response operation) to latent representations.
After the encoder part processing the history data of 14 days, these hidden states pass to the decoder part.
The decoder using GRU and three full connected layer, which accept the hidden state from encoder as initial state and the expected response operation as input, while the output is the prediction of the infection number at next time unit.
Eventually, the local city DT would obtain the infection trend's prediction result in the next 7 days.

\subsection{Federated Training Process: Collaborative City Digital Twin}
\begin{figure}
	\centering
	\includegraphics[width=.48\textwidth]{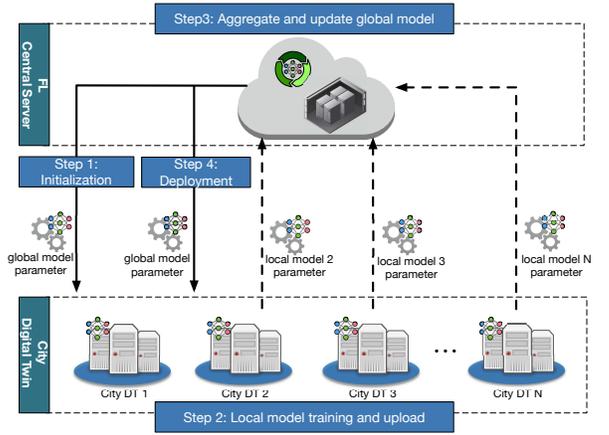}
	\caption{Illustration of the Proposed Federated Training Process}
	\label{fig:flTraining}
\end{figure}
The federated training process aims to provide a collaborative training protocol without violating the privacy rules of each City DT, and the whole training diagram is depicted in Fig.~\ref{fig:flTraining}.
This training process includes following steps:

\begin{enumerate}[(1)]
\item In the beginning, the FL central server trains a global model by pubic data source or voluntary data set from a single DT as pre-training, then open the platform for collaborators to join the training task.
Besides, when a new city DT enters the FL paradigm, the FL central server starts the initialization step again and uses the global model of the last iteration as the new global model.
\item The FL central server distributes the global model to all existing City DTs, and each of them trains the model by the latest local data set to generate the local updates.
\item Each city DT uploads the regional updates to the FL central server for the aggregation step.
\item The FL central server aggregates all the local updates by the aggregation algorithm to generate an updated global model and distributes this model to each city DTs.
This iteration repeats several times until the global model achieves the expected performance.  
\end{enumerate}

Finally, each city DT can always obtain the latest model with another city DT's local updates.
That is, our proposed FL platform can provide an enhanced model for the prediction of COVID-19 infection trend under different response plan, by which each single city DT could determine the response based on crowd-sourcing intelligence without sacrificing privacy.

\begin{figure}
	\centering
	\includegraphics[width=0.4\textwidth]{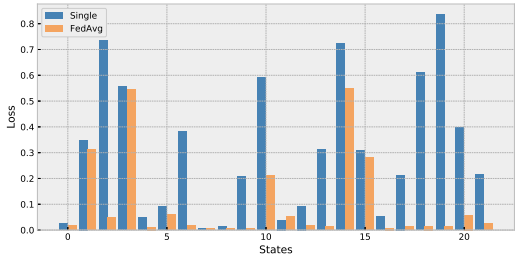}
	\setlength{\abovecaptionskip}{-0.1cm}
	\caption{Loss Comparison of Federated Solution and Non-federated Solution.}
	\label{fig:loss}
\end{figure}

\begin{figure*}[t]
	\centering
	\includegraphics[width=1\textwidth]{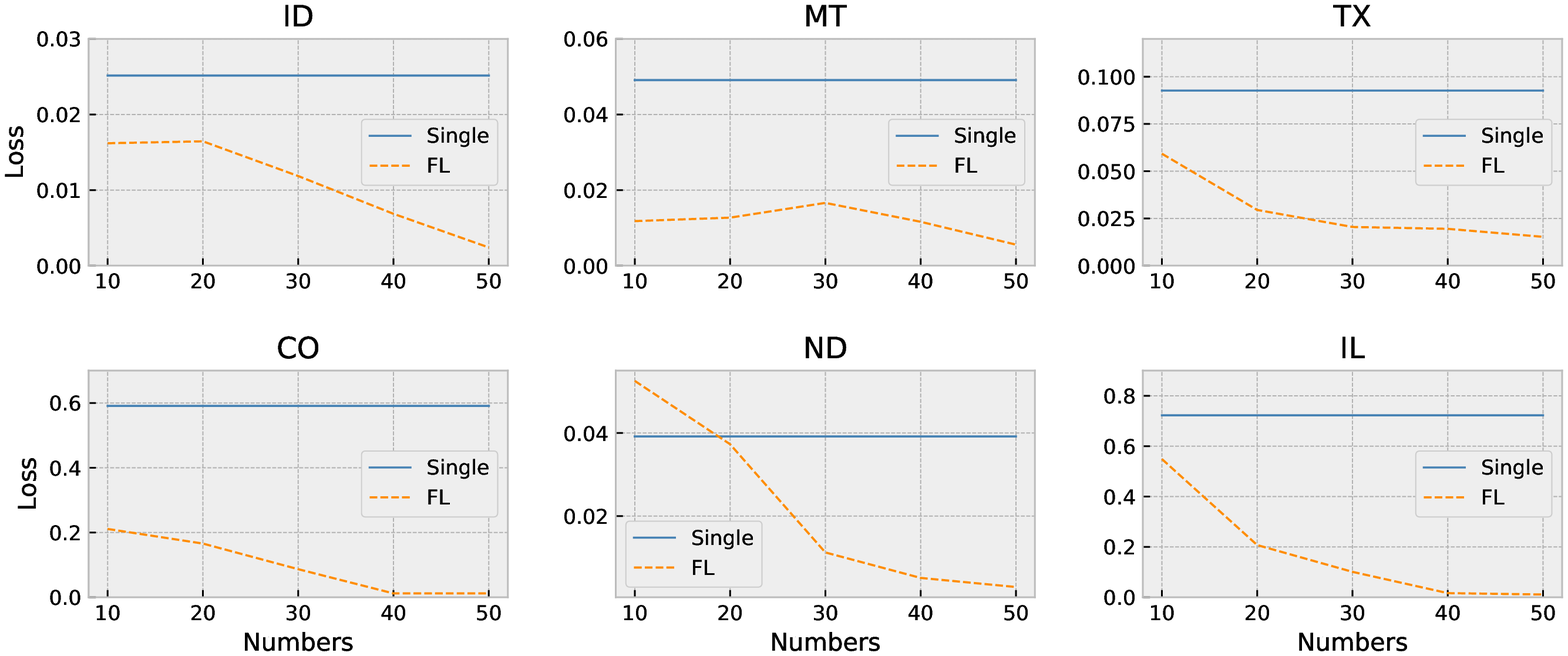}
	\setlength{\abovecaptionskip}{-0.3cm}
	\caption{The prediction error of Federated Learning model with Different Client Numbers}
	\label{fig:fl}
\end{figure*}

\section{Simulation}
\label{simulation}
This section validates our proposed FL based collaborative framework for multiple city DTs through extensive experiments.
First, we give a detailed description of the applied dataset and experiment settings of the framework.
Then, multiple aspects of experiment results, comparisons and analysis are provided, especially the comparisons between non-federated methods and our proposed method in terms of prediction accuracy.

\subsection{Datasets and Experimental Settings}
\textbf{Dataset Description}:
We used the covid tracking project dataset\footnote{https://www.dolthub.com/repositories/Liquidata/corona-virus-state-action} and the Coronavirus (COVID-19) State Actions Dataset\footnote{https://covidtracking.com/data} to conduct experiments. 
The COVID tracking project dataset contains each state's daily epidemic data from the first case in the United States from January, 2020 to July, 2020, including information on the number of nucleic acid tests, the number of confirmed cases, and the number of severe cases. 
The Coronavirus (COVID-19) State Actions Dataset mainly contains specific information about the policies adopted by various states in the United States during the epidemic, including the policy's name, the start time, and the end time.
In this paper, we also need to consider other external factors, such as the change in the number of infections in each state in a time zone, the implementation of various response operations, etc. 
Therefore, we combine the two data sets for our model training.
In our combined dataset, 
\begin{itemize}
 \item $State\ name$ indicates the state where the current data is located.
 \item $Date$ represents the current data time.
 \item $Data-(t-13, \cdots, t+7)$ contains a total of 14 days of epidemic data starting at time $t$ (13 days forward and 7 days backward), and each day’s data is determined by the number of people diagnosed and the current response operation. 
\end{itemize}

A one-hot vector represents every response operation, if it is selected, it is represented as 1; and if not selected, it would be represented as 0.
To eliminate the epidemic data's volatility and highlight the inherent time trend and periodicity, we use the moving average method to calculate the moving average of the number of infections for each day, and the moving window is set to 7.

\textbf{Experiments settings and Parameters}
We conduct all the experiments on the PyTorch platform, and use PySyft~\cite{DBLP:journals/corr/abs-1811-04017} framework for the implementation of our FL protocol.
For the baseline model for local DT training, the encoder is with three-layer GRU and the decoder is with three fully connected layer and a three-layer GRU.
We set the learning rates of the neural networks as $0.005$, mini-batch size as 60, and evaluate the accuracy of prediction results with Mean Square Error (MSE), which is defined as follows:
$$
\mathrm{MSE}=\frac{1}{n} \sum_{i=1}^{n}\left(y_{i}-\hat{y}_{p}\right)^{2}
$$

\subsection{Experiment results}

In Fig.~\ref{fig:loss}, we show the comparison results of our federated solution and non-federated solution.
It shows the difference of the loss value of both of the two models in 22 states, each using the individual training method and federated learning method.
For our federated learning model, FedAvg is adopted as the aggregation algorithm, due to the considerations of saving communication resource. 
The selected response operation includes domestic travel limitations, gatherings limits and stay at home, nonessential business closures, reopening plans and task forces, statewide mask policy.
As shown in Fig.~\ref{fig:loss}, in most states, the loss of epidemic prediction of FL training is much less than that of single DT models. 
This indicates that through FL training, each city DT can exchange the epidemic information of their own city to improve their DT model, so that the prediction performance for COVID-19 trend under selected response operations maintains at a relatively high value.

Fig.~\ref{fig:fl} shows the loss rate of the city DT model from six states (ID: Idaho, MT: Montana, TX: Texas, CO:Colorado, ND: North Dakota, IL: Illinois) with the change of the number of FL clients (collaborators). 
The steel-blue line of each city DT in Fig.~\ref{fig:fl} represents the loss value of the prediction result based on single DT model that only rely on its own dataset, so that it will not change with the number of participating FL members. 
The dark-orange curve shows the loss value of the city DT model that training in the FL environment.
When the number of people participating in the FL platform is increasing, it can be observed the performance has been gradually enhanced, since the information obtained by DT in each city has gradually accumulated.
Under such conditions, the prediction results have become more accurate. 
Thus, the experimental results show that increasing the number of FL clients can effectively improve the training effect and reduce the prediction error of each single city DT.

\section{Conclusion}
\label{conclusion}
In this paper, we propose a novel collaborative city digital twin framework based on federated learning techniques for COVID-19 response plan management, with Seq2seq structure to better capture temporal contexts in historical infection data. 
The proposed framework shows that the collaborative learning of multiple city digital twin is beneficial for alleviating the data sparsity challenge, especially when the novel coronavirus management needs more data sources and highly strict privacy protection requirements.
Besides, the experiments validate our approach achieves better performance on the real dataset.
In the future, we aim to involve more data sources (e.g., movement of people, seasonal changes, temperature, and humidity) to improve the performance of the proposed framework.

\ifCLASSOPTIONcaptionsoff
  \newpage
\fi

\bibliography{ref}

\begin{thebibliography}{10}

\bibitem{WHO}
WHO.
\newblock Coronavirus disease (covid-19) pandemic.
\newblock Website, 2020.
\newblock
  \url{https://www.who.int/emergencies/diseases/novel-coronavirus-2019}.

\bibitem{DBLP:journals/jms/CroattiGMR20}
Angelo Croatti, Matteo Gabellini, Sara Montagna, and Alessandro Ricci.
\newblock On the integration of agents and digital twins in healthcare.
\newblock {\em J. Medical Syst.}, 44(9):161, 2020.

\bibitem{DBLP:books/sp/20/BagariaLBAVE20}
Namrata Bagaria, Fedwa Laamarti, Hawazin~Faiz Badawi, Amani Albraikan, Roberto
  Alejandro~Martinez Velazquez, and Abdulmotaleb El{-}Saddik.
\newblock Health 4.0: Digital twins for health and well-being.
\newblock In Abdulmotaleb El{-}Saddik, M.~Shamim Hossain, and Burak Kantarci,
  editors, {\em Connected Health in Smart Cities}, pages 143--152. Springer,
  2020.

\bibitem{fan2019disaster}
Chao Fan, Cheng Zhang, Alex Yahja, and Ali Mostafavi.
\newblock Disaster city digital twin: A vision for integrating artificial and
  human intelligence for disaster management.
\newblock {\em International Journal of Information Management}, page 102049,
  2019.

\bibitem{DBLP:conf/aistats/McMahanMRHA17}
Brendan McMahan, Eider Moore, Daniel Ramage, Seth Hampson, and
  Blaise~Ag{\"{u}}era y~Arcas.
\newblock Communication-efficient learning of deep networks from decentralized
  data.
\newblock In {\em Proceedings of the 20th International Conference on
  Artificial Intelligence and Statistics, {AISTATS} 2017}, pages 1273--1282,
  2017.

\bibitem{DBLP:conf/aaai/WangCM19}
Lijing Wang, Jiangzhuo Chen, and Madhav Marathe.
\newblock {DEFSI:} deep learning based epidemic forecasting with synthetic
  information.
\newblock In {\em The Thirty-Third {AAAI} Conference on Artificial
  Intelligence, {AAAI} 2019, The Thirty-First Innovative Applications of
  Artificial Intelligence Conference, {IAAI} 2019, The Ninth {AAAI} Symposium
  on Educational Advances in Artificial Intelligence, {EAAI} 2019, Honolulu,
  Hawaii, USA, January 27 - February 1, 2019}, pages 9607--9612. {AAAI} Press,
  2019.

\bibitem{DBLP:conf/icdm/ZhaoCCWLR15}
Liang Zhao, Jiangzhuo Chen, Feng Chen, Wei Wang, Chang{-}Tien Lu, and Naren
  Ramakrishnan.
\newblock Simnest: Social media nested epidemic simulation via online
  semi-supervised deep learning.
\newblock In Charu~C. Aggarwal, Zhi{-}Hua Zhou, Alexander Tuzhilin, Hui Xiong,
  and Xindong Wu, editors, {\em 2015 {IEEE} International Conference on Data
  Mining, {ICDM} 2015, Atlantic City, NJ, USA, November 14-17, 2015}, pages
  639--648. {IEEE} Computer Society, 2015.

\bibitem{DBLP:conf/webi/ShiZBQ019}
Benyun Shi, Jianan Zhong, Qing Bao, Hongjun Qiu, and Jiming Liu.
\newblock Epirep: Learning node representations through epidemic dynamics on
  networks.
\newblock In Payam~M. Barnaghi, Georg Gottlob, Yannis Manolopoulos, Theodoros
  Tzouramanis, and Athena Vakali, editors, {\em 2019 {IEEE/WIC/ACM}
  International Conference on Web Intelligence, {WI} 2019, Thessaloniki,
  Greece, October 14-17, 2019}, pages 486--492. {ACM}, 2019.

\bibitem{lu2019improved}
Fred~S Lu, Mohammad~W Hattab, Cesar~Leonardo Clemente, Matthew Biggerstaff, and
  Mauricio Santillana.
\newblock Improved state-level influenza nowcasting in the united states
  leveraging internet-based data and network approaches.
\newblock {\em Nature communications}, 10(1):1--10, 2019.

\bibitem{DBLP:conf/www/ZouLC18}
Bin Zou, Vasileios Lampos, and Ingemar~J. Cox.
\newblock Multi-task learning improves disease models from web search.
\newblock In Pierre{-}Antoine Champin, Fabien~L. Gandon, Mounia Lalmas, and
  Panagiotis~G. Ipeirotis, editors, {\em Proceedings of the 2018 World Wide Web
  Conference on World Wide Web, {WWW} 2018, Lyon, France, April 23-27, 2018},
  pages 87--96. {ACM}, 2018.

\bibitem{fan2020social}
Chao Fan, Yucheng Jiang, and Ali Mostafavi.
\newblock Social sensing in disaster city digital twin: Integrated
  textual--visual--geo framework for situational awareness during built
  environment disruptions.
\newblock {\em Journal of Management in Engineering}, 36(3):04020002, 2020.

\bibitem{francisco2020smart}
Abigail Francisco, Neda Mohammadi, and John~E Taylor.
\newblock Smart city digital twin--enabled energy management: Toward real-time
  urban building energy benchmarking.
\newblock {\em Journal of Management in Engineering}, 36(2):04019045, 2020.

\bibitem{xue2020lidar}
Fan Xue, Weisheng Lu, Zhe Chen, and Christopher~J Webster.
\newblock From lidar point cloud towards digital twin city: Clustering city
  objects based on gestalt principles.
\newblock {\em ISPRS Journal of Photogrammetry and Remote Sensing},
  167:418--431, 2020.

\bibitem{holstein2018virtual}
WJ~Holstein.
\newblock Virtual singapore-creating an intelligent 3d model to improve
  experiences of residents, business and government, 2018.

\bibitem{dembski2020urban}
Fabian Dembski, Uwe W{\"o}ssner, Mike Letzgus, Michael Ruddat, and Claudia
  Yamu.
\newblock Urban digital twins for smart cities and citizens: The case study of
  herrenberg, germany.
\newblock {\em Sustainability}, 12(6):2307, 2020.

\bibitem{DBLP:conf/ssci/MohammadiT17}
Neda Mohammadi and John~E. Taylor.
\newblock Smart city digital twins.
\newblock In {\em 2017 {IEEE} Symposium Series on Computational Intelligence,
  {SSCI} 2017, Honolulu, HI, USA, November 27 - Dec. 1, 2017}, pages 1--5.
  {IEEE}, 2017.

\bibitem{DBLP:conf/nips/SutskeverVL14}
Ilya Sutskever, Oriol Vinyals, and Quoc~V. Le.
\newblock Sequence to sequence learning with neural networks.
\newblock In Zoubin Ghahramani, Max Welling, Corinna Cortes, Neil~D. Lawrence,
  and Kilian~Q. Weinberger, editors, {\em Advances in Neural Information
  Processing Systems 27: Annual Conference on Neural Information Processing
  Systems 2014, December 8-13 2014, Montreal, Quebec, Canada}, pages
  3104--3112, 2014.

\bibitem{DBLP:journals/corr/abs-1811-04017}
Theo Ryffel, Andrew Trask, Morten Dahl, Bobby Wagner, Jason Mancuso, Daniel
  Rueckert, and Jonathan Passerat{-}Palmbach.
\newblock A generic framework for privacy preserving deep learning.
\newblock {\em CoRR}, abs/1811.04017, 2018.

\end{thebibliography}
\bibliographystyle{unsrt}

\end{document}